\documentclass[conference]{IEEEtran}
\makeatletter
\def\endthebibliography{%
	\def\@noitemerr{\@latex@warning{Empty `thebibliography' environment}}%
	\endlist
}
\makeatother
\IEEEoverridecommandlockouts
\usepackage{cite}
\usepackage{amsmath,amssymb,amsfonts}
\usepackage{algorithm,algpseudocode}
\usepackage{graphicx}
\usepackage{textcomp}
\usepackage{xcolor}
\usepackage{multirow}
\def\BibTeX{{\rm B\kern-.05em{\sc i\kern-.025em b}\kern-.08em
    T\kern-.1667em\lower.7ex\hbox{E}\kern-.125emX}}
\begin{document}

\title{A new node-shift encoding representation for the travelling salesman problem\\
}

\author{\IEEEauthorblockN{Menouar Boulif}
\IEEEauthorblockA{\textit{LIMOSE, Department of computer sciences} \\
\textit{University M'hamed Bougara of Boumerdes}\\
Boumerdes, Algeria \\
boumen7@gmail.com, m.boulif@univ-boumerdes.dz}
\and
\IEEEauthorblockN{Aghiles Gharbi}
\IEEEauthorblockA{\textit{Department of computer sciences} \\
	\textit{University M'hamed Bougara of Boumerdes}\\
	Boumerdes, Algeria\\
	a.gharbi@univ-boumerdes.dz
	}
}

\maketitle

\begin{abstract}
This paper presents a new genetic algorithm encoding representation to solve the travelling salesman problem. To assess the performance of the proposed chromosome structure, we compare
it with state-of-the-art encoding representations. For that purpose, we use 14 benchmarks of different sizes taken from TSPLIB. Finally, after conducting the experimental study, we report the obtained results and draw our conclusion.
\end{abstract}


\section{Introduction}
The travelling salesperson (or salesman) problem (TSP) is one of the oldest combinatorial problems that attracted the attention of notorious scientists \cite{cook2011pursuit}. In fact, there is evidence that quite related problems have been found through ancient manuscripts (especially, in chess game related works such as the knight's tour problem) in the Islamic civilization \cite{murray1913history}. However, according to MM. Flood, the now established TSP form is due to Whitney \cite{flood1956traveling}. TSP can be stated as follows: given a set of cities, TSP aims to find the shortest route to visit each one of them exactly once, and then return to the starting point. \\
Despite its long history, TSP is still among the most challenging NP-complete problems of combinatorial optimization. For this reason it is usually used to assess the performances of new solving approaches \cite{merz2001memetic}.\\
Due to its hardness, many research works tackle TSP by using metaheuristics in which the genetic algorithm (GA) appears to be among the most compatible approach to the TSP landscape \cite{merz2001memetic}. One of the most important constituents of GA that causes it to win or to fail in fulfilling its optimization duty, is the encoding representation. Indeed, the chromosomal structure is the eye with which the GA sees through the landscape it prospects\cite{boulif2019hpga}. Furthermore, if the encoding representation is too bad, then one cannot expect from the GA to reach good solution whatever has been the effort paid to devise the genetic operators. Therefore, we think that conducting more research in this direction deserves more attention in tackling every hard combinatorial problem, and the TSP is not an exception.\\
To contribute to these efforts, this paper proposes a node shift encoding representation (NSE) to solve the TSP. We explain some of NSE's details, and then compare the GA that embed it with an exact method and some state of the art encodings.\\
The rest of this paper is organized as follows. In section 2, we give a short description of two existing encoding representations to solve the TSP. In section 3, we present the NSE representation. Section 4 gives a formulation for the TSP to be used by the exact solving to be considered in the experimental study. Section 5 presents the results of the comparative study we conducted to assess the NSE performances. Finally, we draw our conclusion and present some future axes of research.       

\section{Some existing representations}
Since the first application of the GA on the TSP, there has been several encoding representations. Following are some of the existing approaches:

\subsection{Path encoding}
\begin{figure}[htbp]
	\includegraphics[width=\linewidth]{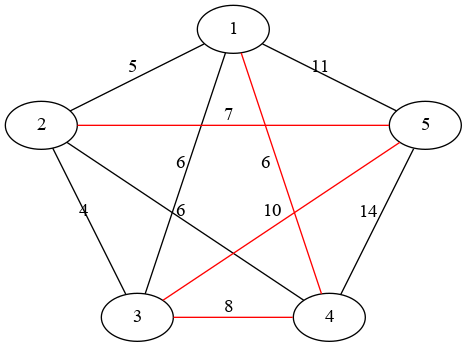}
	\caption{Five cities with corresponding travel costs, and a possible trip (in red, see the electronic version).}
	\label{fig:graphExample}
\end{figure}

Path representation (PR) is by far the most used encoding in the literature \cite{riazi2019genetic}. PR uses a vector of length $n$ that encodes the cities in the order they are visited. For example, a tour passing through five cities, say in order by $1$, $4$, $3$, $5$, $2$ and finally return to the first, can be represented by $(1$ --- $4$ --- $3$ --- $5$ --- $2)$. Notice that the closing route is straightforward, and thus it is omitted.\\
As we shall see in section \ref{secFormulation}, this situation can be represented by an intuitive graph (see Fig. \ref{fig:graphExample}). 

\subsection{Double chromosome}
The double chromosome representation was proposed by \cite{riazi2019genetic}. It uses a vector of even length called the guide chromosome which is a sequence of city index pairs that have to be swapped. The swap is done by using a reference tour called the map chromosome. For example, by considering the map chromosome $(1$ --- $4$ --- $3$ --- $5$ --- $2)$ and the guide $(2,3,1,4)$ we obtain the tour $(5$ --- $3$ --- $4$ --- $1$ --- $2)$ by swapping the $2^{nd}$ index with the $3^{rd}$, then the $1^{st}$ with the $4^{th}$.\\       

For a quite exhaustive presentation of the existing representations, we refer the interested reader to \cite{potvin1996genetic, larranaga1999genetic}. 
\section{Formulation}
\label{secFormulation}
TSP can be naturally modelled by a digraph whose vertices are the cities, and there is an arc between two vertices $iff$ there is a route that directly links the corresponding cities. The arcs are weighted by the distance or the cost of the associated route. When the outward and return travel costs are the same for every linked cities, we can use a undirected graph instead. Fig. \ref{fig:graphExample} depicts such a graph for a five cities travel plan.\\      

We can also model TSP with a mathematical programming approach. Following is an integer linear program for the TSP known as the Miller–Tucker–Zemlin (MTZ) formulation \cite{miller1960integer}:  
\begin{center}
	\includegraphics[width=\linewidth]{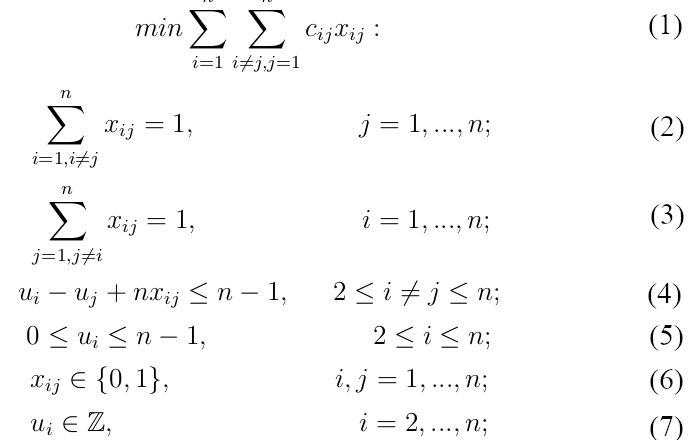}
\end{center}

This formulation considers a set of $n$ cities, $V = \{v_1, v_2, ..., v_n\}$, to be visited by the travelling salesperson. In the objective function ($eq.\text{ }1$), the coefficient $c_{ij}$ denotes how much it will cost to travel directly from $v_i$ to $v_j$. The main decision variable of the model $x_{ij}$ is defined as follows:
	$$
	x_{ij} = \left\{
	\begin{array}{ll}
		1 & \mbox{if the salesperson travels directly from $v_i$ to $v_j$}. \\
		0 & \mbox{Otherwise.}
	\end{array}
	\right.
	$$
Hence, the objective functions minimizes the cost of the overall travel.\\
Equations $2$ and $3$ are coherence constraints insuring that the salesperson will visit every city only once.\\ The second decision variable $u_i$ with its related equations (i.e. $4$ and $5$) are added in order to insure the travel is one big closed tour that goes through all the cities.\\

MTZ is among the most recognized seminal works in its domain \cite{oncan2009comparative, diaby2016advances} due to its compactness. We will use it for assessing the performance of the proposed approach.

\section{Contribution}
The complexity of the TSP triggered a big amount of works that use approximate solving approaches. Metaheuristics are such a method that can find optimal or near optimal solutions in a reasonable period of time. Genetic algorithms are among the approximate approaches that have proven ability to solve hard combinatorial problems. For further details on the GA method the interested reader can use \cite{goldberg1989genetic}. In what follows, we describe the new encoding representation we will use to implement our GA.   
\subsection{Node shift encoding representation}
\begin{figure*}[htbp]
	\centerline{\includegraphics[width=\linewidth]{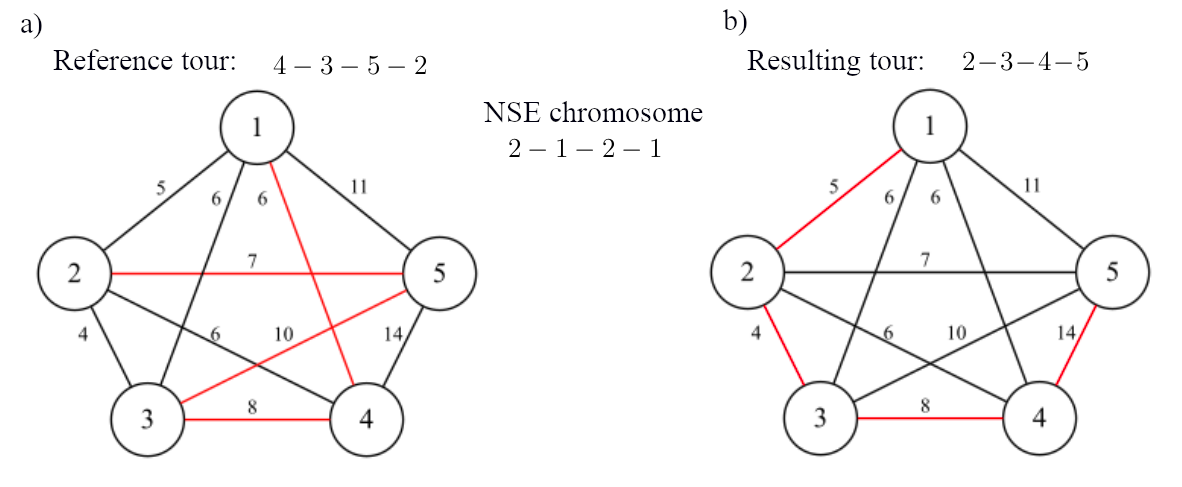}}
	\caption{Getting the NSE solution by combining the reference tour with the NSE chromosome.}
	\label{fig:NSEvectorEx}
\end{figure*}

The Node Shift Encoding (NSE) belongs to the ordinal representations class. NSE uses a reference tour which is a sequence of city indexes, and encodes upon it the number of moves an index has to achieve to reach its new position. Given the position of a city index in the reference tour, the moves are done from left to right. If the moving index reaches the end of the sequence, it continues from the beginning, thus making the moves to act in a circular manner. In another hand, the index of the first city in the reference solution is always put in the first place, and hence it can be hidden. NSE representation uses a vector of length $n-1$ that defines the number of moves for each index of the reference sequence to be done in a \textit{sequential} order. An example will make it more clear. Given the reference tour of Fig. \ref{fig:NSEvectorEx} a), which by adding the hidden city equivalent to $(1,4,3,5,2)$, the NSE encoding representation $(2,1,2,1)$ informs us that: 
\begin{itemize}
	\item The city index of $v_4$ is moved forward by two positions yielding the index sequence $(1,3,5,4,2)$.
	\item Then, the city index of $v_3$ is moved forward by one position yielding the index sequence $(1,5,3,4,2)$. $v_3$ has been chosen because the moves defined in the NSE chromosome are always associated to the reference tour.
	\item Then, the city index of $v_5$ is moved forward by two position yielding the index sequence $(1,3,4,5,2)$.
	\item Then, the city index of $v_2$ is moved forward by one position. $v_2$ being the last index, it performs its shift from the beginning yielding the index sequence $(1,2,3,4,5)$.
\end{itemize}
Hence, the NSE chromosome $(2,1,2,1)$ is the encoding representation of the tour $(1,2,3,4,5)$ (see Fig. \ref{fig:NSEvectorEx} b)).\\ 

\begin{figure}[htbp]
	\includegraphics[width=\linewidth]{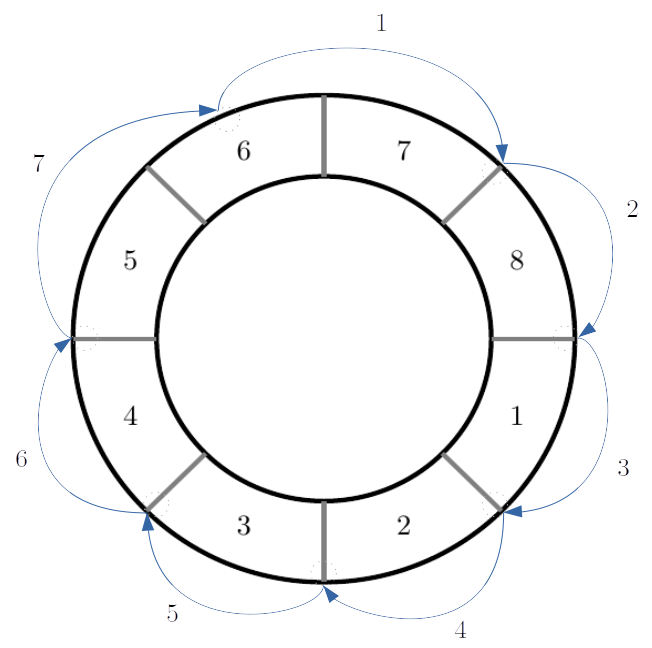}
	\caption{In a length 8 sequence, shifting index 6 for 7 steps brings it to the starting place.}
	\label{fig:NSEboundExpl}
\end{figure}

It is worth mentioning that every allele of the NSE chromosome can be bounded by the interval $[0,n-2]$, where $n$ is the length of the tour (i.e. the number of cities). Indeed, since the moves are done in a circular manner, a number of shifts $ns$ that exceeds $n$ will be actually doing $ns \mod n-1$ moves, because every $n-1$ of them bring the shifted index to the starting point (see Fig. \ref{fig:NSEboundExpl}). \\

Finally, decoding an NSE chromosome to get the associated tour can be achieved by using Algorithm \ref{NSEdecodingAlgo}. The NSE decoding algorithm accepts as input a reference tour $refTour$ encoded by a path representation, and an NSE vector $chromo$. Then, it moves the $refTour$ indexes having a positive shift number in $chromo$ in a sequential and circular manner to finally get the solution $tour$ with a path representation.\\

\begin{algorithm}
\caption{NSE decoding procedure}
\label{NSEdecodingAlgo}
\begin{algorithmic}[1]
	\State \bf{Input:} $refTour, chromo$
	\State \bf{Output:} $tour$
	\State $len \leftarrow length(refTour)$
	\State $vRank \leftarrow {1:len} $
	\For{$i\leftarrow 2\text{ }\bf{to}\text{ }$$len$}
	\State $oldRank \leftarrow vRank[i]$
	\State $newRank \leftarrow vRank[i] + chromo[i - 1]$
	\If{$newRank > len$}
	\State $newRank \leftarrow newRank-len+1$
	\EndIf
	\If{$newRank > oldRank$}
	\For{$j\leftarrow 1\text{ }\bf{to}\text{ }$$len$}
	\If{$vRank[j] \leq newRank$ \bf{and} \newline \hspace*{4em} $vRank[j] \geq oldRank$}
	\State $vRank[j] \leftarrow vRank[j]-1 $
	\EndIf
	\EndFor
	\Else
	\For{$j\leftarrow 1\text{ }\bf{to}\text{ }$$len$}
	\If{$vRank[j] < oldRank $ \bf{and} \newline 
	\hspace*{4em}$\text{ }vRank[j] \geq newRank$}
	\State $vRank[j] \leftarrow vRank[j]+1 $
	\EndIf
	\EndFor
	\EndIf
	\State $vRank[i] \leftarrow newRank$
	\EndFor
	\For{$i\leftarrow 1\text{ }\bf{to}\text{ }$$len$}
	\State $tour[vRank[i]] \leftarrow refTour[i]$
	\EndFor
\end{algorithmic}
\end{algorithm}


\section{Experimental study}
In order to assess NSE performances, we compare it to the path representation (PR) and the double chromosome (DC) encoding. Each one of these encodings has been embedded on the basic elitist GA of the R package gramEvol \cite{Farzad2016gramevol} that uses simple operators such as one point crossover and simple mutation. For each GA of these three, we use two variants for the initial population: the first uses only random individuals, whereas the second injects the best solution found by the Nearest Neighbourhood (NN) heuristic \cite{gutin2002traveling}. Hence, in what follows, we refer to the six so constructed variants by NSE-RAND, NSE-NN, PR-RAND, PR-NN, DC-RAND and DC-NN. Besides, by using the integer linear program presented in Section \ref{secFormulation} along with the Rglpk tool \cite{Theussl2019rglpk}, we got an exact method. We shall denote it by GLPK.\\        
We implemented the six plus one methods in R 3.6.3 \cite{rcoreteam2020}, and run them on a machine equipped with an intel core i5-7200U, 2.5-3.1 GHz CPU, and 4Go of RAM.\\
We took 14 benchmarks from \cite{reinelt1991tsplib} (see Table \ref{benchmarks}). We divided them into three classes according to their size. 

\begin{table}
	\centering
	\begin{tabular}{c l c c }
		\hline
		\multirow{2}{*}{\textbf{Problem number}} & \multirow{2}{*}{\textbf{Name}} & \multirow{2}{*}{\textbf{Number of cities}} & \multirow{2}{*}{\textbf{Class}} \\
		& & \\
		\hline
		1 & eil51 & 51 & \multirow{5}{*}{1}\\  
		2 & berlin52 & 52  & \\
		3 & st70 & 70   & \\
		4 & eil76 & 76   & \\
		5 & rat99 & 99  & \\
		\hline
		6 & kroB100 & 100   & \multirow{5}{*}{2} \\
		7 & kroA100 & 100   & \\
		8 & rd100 & 100  & \\
		9 & eil101 & 101  & \\
		10 & lin105 & 105  & \\
		\hline
		11 & ch130 & 130   & \multirow{4}{*}{3} \\
		12 & ch150 & 150   & \\
		13 & d198 & 198  & \\
		14 & kroA200 & 200 & \\
		\hline
	\end{tabular}
	\caption{Benchmarks.}
	\label{benchmarks}
\end{table}

Before starting the tests, we looked for the best parameters for each GA variant. 
We did that by considering the largest benchmark from each class and tested it with all the parameter combinations within the following values:
\begin{itemize}
	\item Population size (50, 100, 500, 1000);
	\item Number of iterations (100, 500, 1000, 2000);
	\item Mutation chance (0.01, 0.03, 0.05, 0.1);
\end{itemize}

We took the best combination of parameters and adopt it to run the six approximate methods thirty times and reported the best result for each variant. Table \ref{results} gives the obtained best solutions in terms of tour cost.

\begin{table*}
	\centering
	\begin{tabular}{ l | l l | l l | l l | l l l l}
		\hline
		\multirow{2}{*}{\textbf{Benchmark}} & \multicolumn{2}{c}{\textbf{PR}} & \multicolumn{2}{c}{\textbf{DC}} & \multicolumn{2}{c}{\textbf{NSE}} & \multirow{2}{*}{\textbf{GLPK}} & \multirow{2}{*}{\textbf{Optimal}}\\
		& \textbf{NN} & \textbf{ RND} & \textbf{NN} & \textbf{ RND} & \textbf{NN} & \textbf{ RND} &  &  \\
		\hline
		eil51 & 549 & 577 & 540 & 529 & 440 & \textbf{436} & 436 & 426 \\
		berlin52 & 10475 & 9590 &  9411 & 9167 & 8225 & \textbf{7824} & 7695 & 7542 \\
		st70 & 1184 & 1119 & 1065 & 1043 & \textbf{702} & 705 & 773 & 675 \\
		eil76 & 847 & 848 & 851 & 821 & \textbf{574} & 577 & 583 & 538 \\
		rat99 & 1750 & 1707 & 2043 & 2662 & 1561 & \textbf{1433} & 1337 & 1211\\
		kroB100 & 44414 & 46027 & 55280 & 54802 & 27073 & \textbf{25630} & 29130 & 22141\\
		kroA100 & 50225 & 47823 & 55891 & 53842 & \textbf{24671} & 24906 & 24729 & 21282 \\
		rd100 & 16726 & 17066 & 18788 & 17447 & \textbf{9147} & 9711 & 9226 & 7910 \\
		eil101 & 1106 & 1084 & 1260 & 1231 & 725 & \textbf{721} & 666 & 629\\
		lin105 & 22440 & 23605 & 33757 & 37229 & 19362 & \textbf{19139} & 21337 & 14379 \\
		ch130 & 15194 & 15573 & 18246 & 18702 & \textbf{8087} & 8421 & 7679 & 6110 \\
		ch150 & 20748 & 18350 & 23999 & 24082 & \textbf{9995} & 10201 & 7857 & 6528 \\
		d198 & \textbf{22329} & 23788 & 32124 & 70324 & 28069 & 28024 & 27154 & 15780 \\
		kroA200 & 125014 & 123719 & 167184 & 166162 & \textbf{57678} & 58532 & 60907 & 29368 \\
		\hline
    \end{tabular}
	\caption{Best results for 14 benchmarks, 30 runs for each.}
	\label{results}
\end{table*}

\begin{figure}[htbp]
	\includegraphics[width=\linewidth]{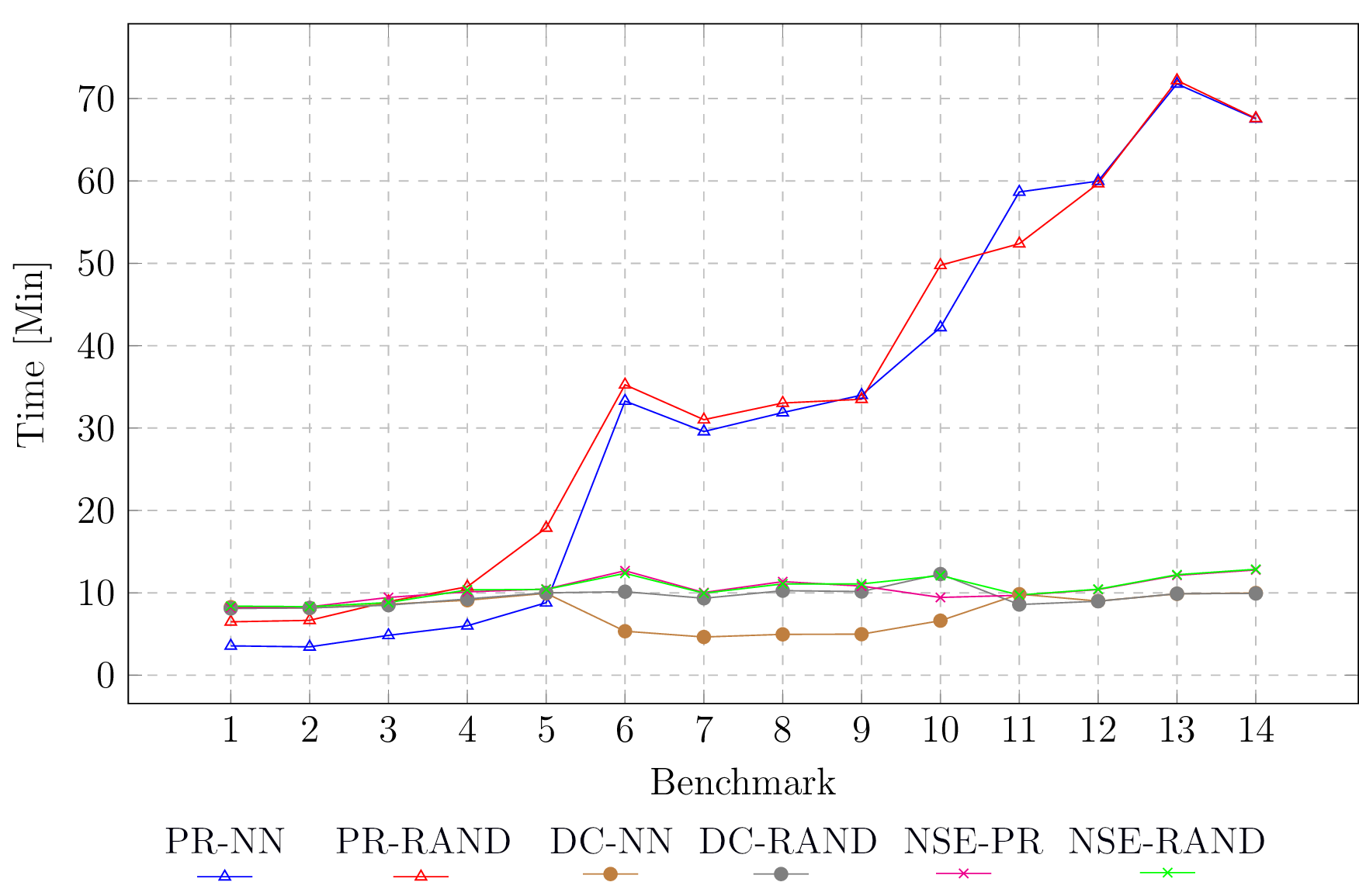}
	\caption{Mean runtime for the 30 runs.}
	\label{fig:meanRunTime}
\end{figure}

In addition to the best results of the six approximate methods, Table \ref{results} presents the best tour cost found by the exact method, GLPK, in 4 hours. The best results when comparing the approximate methods are boldfaced. The value of the optimal solution taken from \cite{reinelt1991tsplib} is reported in the last column, Optimal. The mean runtime of the 30 runs are depicted in Fig. \ref{fig:meanRunTime}.\\

\begin{figure*}[htbp]
	\includegraphics[width=\linewidth]{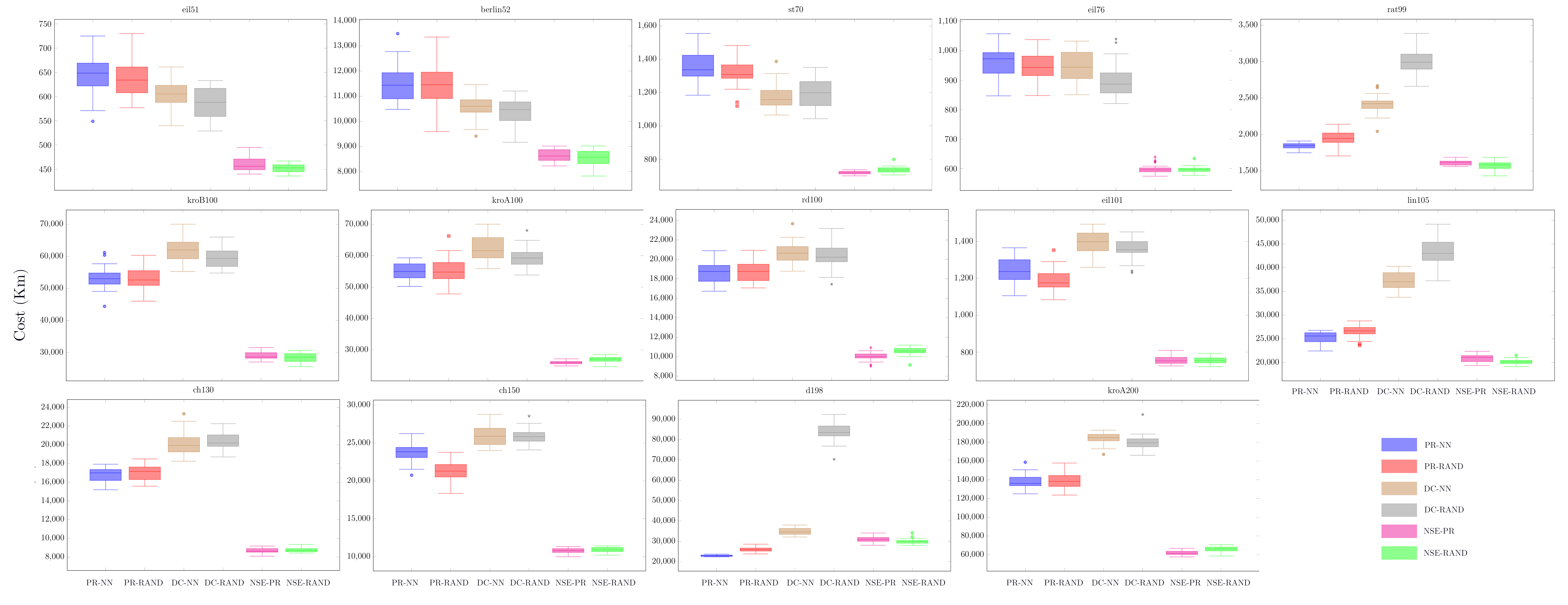}
	\caption{Boxplots of results for 30 runs.}
	\label{fig:boxCompil}
\end{figure*}

By comparing the performances of the approximate methods, we notice that NSE gives the best performances in all but one benchmark. Furthermore, in eight out of fourteen benchmarks, NSE gives better results than GLPK. NN heuristic doesn't seem to procure a great help to NSE in reaching better solutions especially when the size gets bigger. This can be further observed by the position and the sizes of the NSE boxplots in Fig. \ref{fig:boxCompil}.
In another hand, concerning the running time, we notice that the mean runtimes for DC and NSE are close to each other. For PR, the running time became extremely huge for big instances (see Fig. \ref{fig:meanRunTime}).\\
Moreover, none of the obtained solutions were optimal, and the tiny form of the NSE boxplots, especially when the benchmark size gets bigger, informs us that the new encoding is easily trapped in local optima, and hence suggests the need for more sophisticated mutation or other diversification mechanism.\\      
\section{Conclusion}
We proposed a Node Shift Encoding (NSE) which is a new encoding representation to solve the Travelling Salesperson Problem (TSP) with the genetic algorithm. We conducted a comparative study to assess the performances of NSE in front of the path representation (PR), which is the most used encoding in the literature, and the double chromosome (DC) representation. The obtained results reveal that the new encoding is promising.The experimental study showed also that using the nearest neighbour heuristic to have some starting solutions inserted in the initial population doesn't procure a clear help to NSE and DC but PR. In addition, the relatively stable performance of NSE suggests it may require additional diversification operators.\\

As future work, since NSE was embedded into a simple GA, we are interested in analysing how it will behave when associated to more conceptually minded operators. Furthermore, applying NSE on other problems closely related to the TSP such as the Vehicle Routing Problem (VRP) and its variants seems to be another promising axis of research.

\bibliographystyle{ieeetran}
\bibliography{references.bib}

\end{document}